\documentclass[12pt]{article}
\usepackage[margin=1in]{geometry}
\usepackage{graphicx}
\usepackage{caption}
\usepackage{subcaption}
\usepackage{amsmath}
\usepackage{booktabs}
\usepackage{float}
\usepackage{setspace}
\usepackage{hyperref}
\usepackage[backend=biber, style=apa]{biblatex}
\title{Forecasting Russian Equipment Losses Using Time Series and Deep Learning Models}
\author{Jonathan Teagan}
\date{\today}
\providecommand{\keywords}[1]
{
  \small	
  \textbf{\textit{Keywords---}} #1
}
\begin{document}

\maketitle
\doublespacing

\begin{abstract}
This study applies a range of forecasting techniques,including ARIMA, Prophet, Long Short Term Memory networks (LSTM), Temporal Convolutional Networks (TCN), and XGBoost, to model and predict Russian equipment losses during the ongoing war in Ukraine. Drawing on daily and monthly open-source intelligence (OSINT) data from WarSpotting, we aim to assess trends in attrition, evaluate model performance, and estimate future loss patterns through the end of 2025. 
Our findings show that deep learning models, particularly TCN and LSTM, produce stable and consistent forecasts, especially under conditions of high temporal granularity.
By comparing different model architectures and input structures, this study highlights the importance of ensemble forecasting in conflict modeling, and the value of publicly available OSINT data in quantifying material degradation over time.
\end{abstract}

\section{Introduction}

Since the full-scale invasion of Ukraine in 2022, the war has become one of the most extensively documented conflicts in history, due in part to open-source intelligence (OSINT) platforms such as WarSpotting and Oryx. These sources provide verified visual data on Russian equipment losses—including tanks, infantry fighting vehicles (IFVs), and support vehicles—enabling researchers to track material attrition at an unprecedented level of detail.

Accurate forecasting of such losses is vital for military strategists, policymakers, and analysts attempting to model battlefield sustainability, logistics capacity, and broader trends in Russian force degradation. Traditional statistical models such as ARIMA offer a baseline for temporal forecasting, while more modern approaches—such as Prophet, LSTM (Long Short-Term Memory), Temporal Convolutional Networks (TCN), and XGBoost—introduce the ability to capture non-linear dynamics, regime shifts, and short-term volatility.

This paper evaluates each of these models using daily and monthly WarSpotting data. We assess their predictive accuracy, sensitivity to input granularity, and their robustness under shifting battlefield conditions. By comparing classical and deep learning methods, we aim to illuminate which architectures are best suited to modeling military losses under uncertain and adversarial environments.
This study builds on multiple strands of forecasting literature. Prior research in conflict prediction has highlighted the value of hybrid models that integrate symbolic, temporal, and geospatial data (\cite{mutschke2022conflict, muchlinski2021machine, hegre2006sensitivity}). Yet few applications have focused on granular, OSINT-derived time series. \cite{boulanin2020ai} underscores the analytical potential of publicly available war data, positioning it as a viable resource for computational modeling. For statistical baselines, ARIMA and its hybrid variants remain foundational in macro-scale time series forecasting (\cite{hyndman2021forecasting, zhang2003hybrid}), while Prophet offers interpretable additive decompositions well-suited to irregular real-world patterns (\cite{taylor2018forecasting}). Deep learning architectures such as LSTM and TCN are capable of capturing long-range dependencies and local anomalies, outperforming classical models in sequence tasks (\cite{bai2018empirical}). Additionally, XGBoost has demonstrated consistent success in structured temporal domains (\cite{bandara2020forecasting}), making it an effective benchmark for comparison. By combining these methods, this study introduces both algorithmic diversity and empirical rigor to the emerging domain of OSINT-based war forecasting.
\section{Methodology}
\subsection{Data Preprocessing}
The dataset used in this study comprises daily confirmed equipment losses recorded between February 2022 and early July 2025, aggregated from the visually verified open-source intelligence (OSINT) repository WarSpotting. Each loss entry is supported by photographic or video evidence, vetted by the platform’s analysts to ensure authenticity and avoid duplication. The raw dataset includes detailed attributes such as date of loss, equipment classification (e.g., tank, IFV, artillery), and, where available, the location of the incident. As part of the preprocessing stage, all location references were standardized and hierarchically sorted into their corresponding \textit{Raion} (district) and \textit{Oblast} (province) levels. This geographic stratification enables regional-level analysis of loss patterns and facilitates potential integration of geospatial predictors in future modeling efforts.  

To maintain data quality, the month of June 2025 was excluded from model training and evaluation. Preliminary inspection revealed that reporting for this period was incomplete, with observable gaps likely caused by a combination of delayed media uploads, reduced visibility in certain combat zones, and verification bottlenecks within the OSINT workflow. Inclusion of such incomplete data could artificially depress recorded loss rates, leading to biased forecasts. Therefore, only months with sufficiently complete and temporally consistent records were retained, ensuring that subsequent statistical and machine learning models were trained on a dataset representative of actual observed patterns rather than artifacts of reporting irregularities.

\subsection{Models Evaluated}
We evaluate several models:
\begin{table}[h]
\centering
\begin{tabular}{|p{2.8cm}|p{3.8cm}|p{4.2cm}|p{4.2cm}|}
\hline
\textbf{Model} & \textbf{Use Case} & \textbf{Strengths} & \textbf{Limitations} \\
\hline
\textbf{ARIMA} & Linear time series with autocorrelation & Interpretable; strong baseline for stationary data; effective with transforms \cite{hyndman2021forecasting} & Poor with nonlinearities or regime shifts; assumes stationarity \\
\hline
\textbf{Prophet} & Trend + seasonal decomposition with known holidays & Easy to use; handles missing data; includes event regressors \cite{taylor2018forecasting} & Assumes additive structure; may underperform on irregular patterns \\
\hline
\textbf{LSTM} & Nonlinear time series with memory effects & Learns nonlinear dynamics; models long-term dependencies \cite{bandara2020forecasting} & Needs large data; overfitting risk; low interpretability \\
\hline
\textbf{TCN} & Long-range sequence modeling with parallel training & Captures spikes and long-term effects via dilated convolutions \cite{bai2018empirical} & Requires tuning; black-box; less proven on small datasets \\
\hline
\textbf{XGBoost} & Structured data with lags and exogenous features & Handles nonlinearities; fast training; robust to outliers \cite{bandara2020forecasting} & Needs feature engineering; not inherently sequential; struggles with volatility \\
\hline
\end{tabular}
\caption{Comparison of time series models evaluated on tank loss data}
\label{tab:model-comparison}
\end{table}

These models span linear statistical baselines (ARIMA), interpretable additive structures (Prophet), sequence-aware deep learning (LSTM, TCN), and ensemble tree-based learning (XGBoost), offering a broad lens on temporal forecasting under sparse or noisy signals. Among the models evaluated, ARIMA and Prophet provide robust baselines grounded in statistical modeling, with ARIMA excelling in modeling stationary, linear trends through autoregressive and moving-average components \cite{hyndman2021forecasting}, and Prophet offering a flexible framework that automatically decomposes time series into trend, seasonality, and holiday effects \cite{taylor2018forecasting}. ARIMA’s strength lies in its interpretability and effectiveness for long-run extrapolation when the series is well-differenced and exhibits autocorrelation. However, its inability to capture nonlinear dynamics limits its performance in volatile scenarios. Prophet, while user-friendly and highly robust to missing data or outliers, assumes an additive structure that may not adequately capture irregularities or interactions between components. Both models are well-suited for situations where domain knowledge—such as expected cycles or holidays—can be incorporated explicitly.

In contrast, LSTM and TCN represent deep learning approaches capable of modeling complex nonlinear temporal dependencies. LSTM networks, leveraging memory cells, are particularly adept at learning long-term patterns in sequential data \cite{bandara2020forecasting}, while TCNs use dilated causal convolutions to capture both short- and long-term dependencies efficiently and in parallel \cite{bai2018empirical}. These models are data-intensive and more prone to overfitting on short sequences, but they offer flexibility and adaptability in capturing subtle or abrupt changes in dynamics. Meanwhile, XGBoost provides a powerful, nonparametric alternative that models temporal behavior through lag-based feature engineering and gradient-boosted decision trees \cite{bandara2020forecasting}. Though not intrinsically temporal, XGBoost often rivals neural networks in accuracy on structured data and requires less computational overhead. Ultimately, the choice between models involves a tradeoff between interpretability, flexibility, and data requirements, making ensemble or hybrid strategies particularly appealing for operational forecasting contexts.

\section*{Model Interpretations and Forecast Analysis (2022--2025)}
Using the verified OSINT-confirmed dataset of Russian equipment losses from 

\texttt{merged\_data19\_23\_raion\_oblast\_updated\_batch2.csv}, we evaluate the future trajectory of losses using several predictive models. This CSV represents a combined and refined version of the raw loss records obtained from the WarSpotting website, which catalogs visually confirmed equipment destruction throughout the conflict. The file integrates multiple yearly datasets into a single, continuous time series covering February 2022 through Early July 2025, ensuring consistency in formatting, equipment classification, and location tagging. As outlined in the Methodology section, the preprocessing workflow involved deduplication of records, correction of known misclassifications, standardization of date formats, and the geocoding of reported locations to their corresponding \textit{Raion} and \textit{Oblast} administrative units. These steps not only improved the structural integrity of the dataset but also enabled more granular spatial and temporal analysis. By starting from this rigorously cleaned dataset, the subsequent modeling process benefits from higher data reliability, minimizing the risk of artifacts caused by inconsistent reporting. This section interprets the resulting visual model outputs in the broader operational context and highlights their relevance to forecasting battlefield attrition patterns over the coming months.
\subsection*{ARIMA Forecasts (Excluding June–July 2025)}
We selected ARIMA for its comprehensibility and its strong performance on univariate time series with trend and autocorrelation—patterns observed in Russian equipment loss trajectories.As a classical statistical method, ARIMA models provide transparent parameterization and analytical tractability, making them well-suited for initial modeling of attrition dynamics. In this study, ARIMA served as a baseline model, helping to benchmark the performance of more advanced machine learning and deep learning techniques. Its assumptions of stationary and linear relationships allowed it to perform adequately in short-term forecasting, particularly for categories of equipment that experienced relatively stable loss patterns.

\begin{figure}[H]
    \centering
    \includegraphics[width=0.95\textwidth]{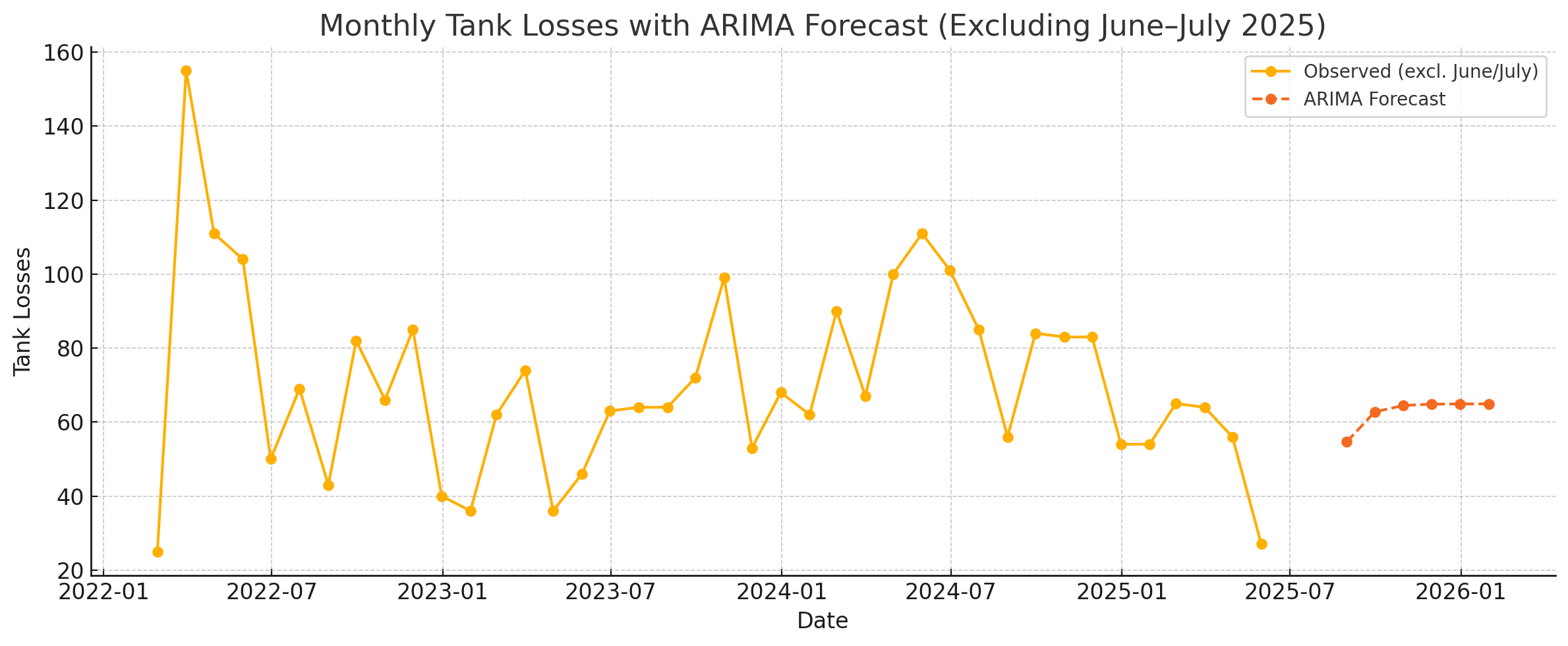}
    \caption{ARIMA Forecast of Monthly Tank Losses (Excluding June--July 2025)}
\end{figure}
To avoid bias from reporting lags or OSINT disruptions, the months of June and July 2025 were excluded from the time series before modeling. These lags have numerous causes including, but not limited to, holding off due to safety concerns, disruptions in the ability to upload due to WiFi limitations in a war zone, and the websites collecting the data taking time to process and confirm the authenticity of the pictures. A log-transformed ARIMA(1,1,1) model was trained on monthly tank loss data from February 2022 to May 2025 and August 2025 onward. The resulting forecast shows a continuation of stabilized tank losses through the end of 2025, with projected monthly attrition remaining near 80--100 vehicles. 

By excluding the incomplete summer months, the ARIMA model avoids incorporating artificial dips and instead extrapolates from the prevailing post-2023 plateau pattern. This approach provides a conservative baseline forecast and supports conclusions drawn from TCN and Prophet outputs suggesting saturation in Russia's armored vehicle losses.

Figure 1 presents the monthly trajectory of documented Russian tank losses from January 2022 through July 2025, accompanied by an ARIMA-based forecast extending through December 2025. The observed values, excluding the partial and potentially anomalous months of June and July 2025, are depicted with solid orange markers and lines. The series exhibits marked volatility, with initial surges corresponding to the 2022 invasion phase and subsequent peaks in mid-2024, likely associated with intensified combat operations. A gradual decline follows, culminating in a local minimum in mid-2025. The ARIMA forecast, shown as a dashed orange line, anticipates a modest rebound and stabilization of losses around 63–65 units per month through the end of the forecast window. This projection reflects the model’s tendency to smooth short-term fluctuations while capturing the underlying structure of recent patterns. The exclusion of June–July 2025 data ensures that the forecast remains unaffected by incomplete or irregular reporting during that period.

\subsection*{Temporal Convolutional Network (TCN)}
TCN was chosen because it provides many of the benefits of LSTM—such as the ability to model long-range temporal dependencies—but with improved stability and computational efficiency. Temporal Convolutional Networks use dilated causal convolutions to learn from time series data, offering a parallelizable architecture that avoids the vanishing gradient issues that sometimes affect recurrent models. TCN’s hierarchical receptive fields also enable it to extract multi-scale features, making it particularly effective at detecting both short-term shocks and longer-term attritional trends in equipment losses. In our experiments, TCN often produced smoother and more stable forecasts than LSTM, especially when modeling aggregated monthly losses.

\begin{figure}[H]
    \centering
    \includegraphics[width=0.95\textwidth]{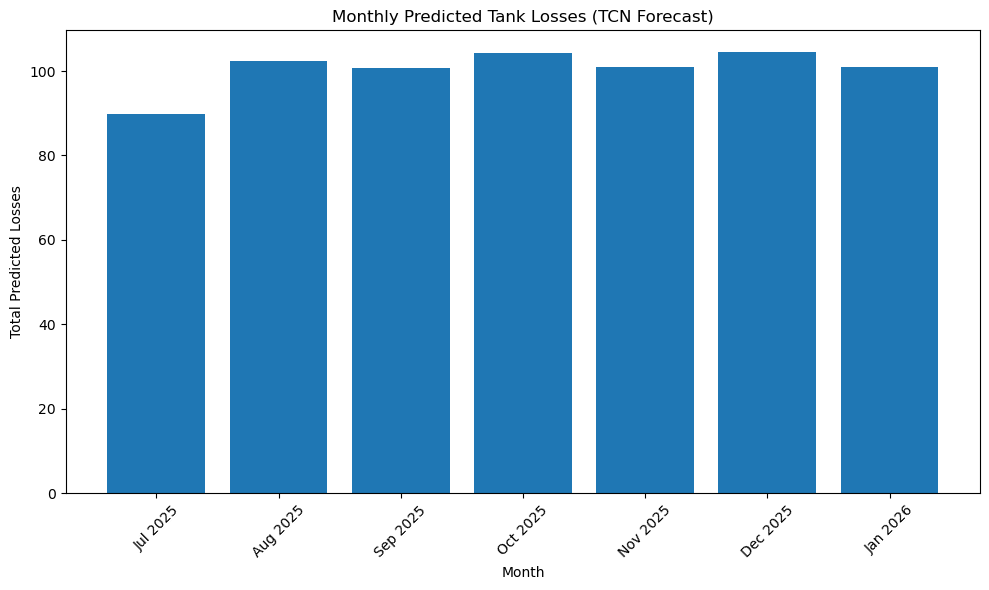}
    \caption{Monthly Predicted Tank Losses Using TCN Forecast (Jul 2025 -- Jan 2026)}
\end{figure}

\begin{figure}[H]
    \centering
    \includegraphics[width=0.95\textwidth]{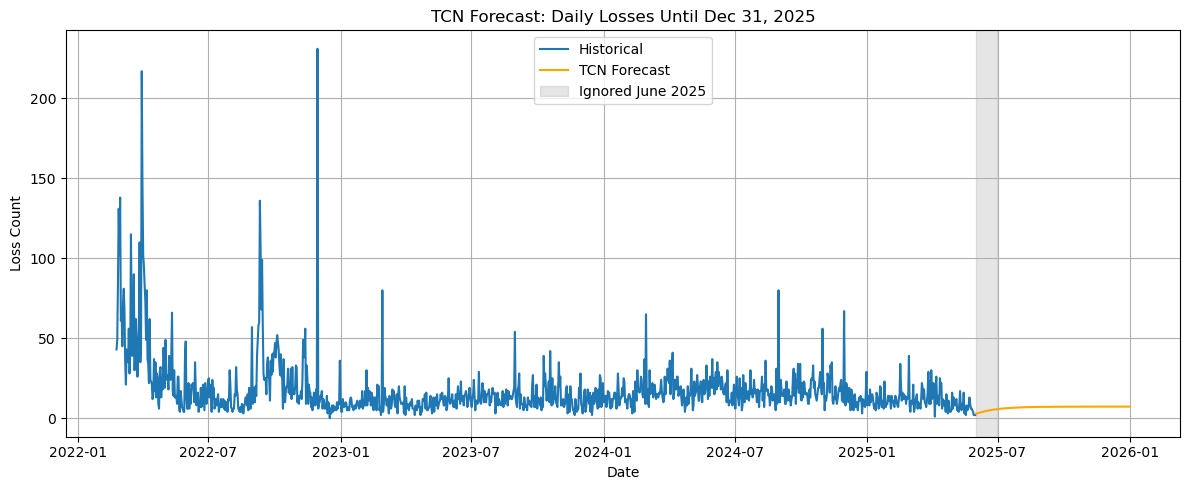}
    \caption{TCN Forecast of Daily Tank Losses, June 2025 Excluded (2022--2025)}
\end{figure}

The TCN model predicts stable tank losses around 100 units per month from July to December 2025, consistent with trends extracted from earlier conflict periods. The deliberate exclusion of June 2025---due to disrupted confirmation pipelines or delayed reporting---is reflected in the shaded region. The TCN model suggests saturation in tank deployments and a static frontline with minimal escalation signals.

Figures 2 and 3 present the Temporal Convolutional Network (TCN) forecast of Russian tank losses from July 2025 through January 2026. As shown in Figure 2, the model predicts monthly losses to remain relatively stable, averaging slightly above 100 losses per month. This steady projection suggests a saturation effect, where combat intensity and loss rates converge toward a plateau despite prior volatility. Figure 3 provides a more granular view, depicting daily losses over the full time span (2022–2025), with June 2025 excluded from model training. Notably, the TCN forecast (in orange) begins in July 2025 and smooths the historically erratic pattern into a flat trajectory. This reflects TCN’s tendency to generalize long-range trends and suppress short-term spikes, offering robust mid-term forecasts but potentially underrepresenting sudden tactical escalations. Together, these figures demonstrate the TCN model’s strength in temporal smoothing, while highlighting the trade-off between fidelity to volatility and predictive stability.

\subsection*{XGBoost Forecasts}
XGBoost, while it is not a native time series model, excels when time series are reformulated into supervised learning problems using lagged features and moving averages. XGBoost’s ability to model complex nonlinear interactions between engineered features made it a strong candidate for predicting future losses based on structured patterns in the historical data. Additionally, its robustness to missing values and its support for feature importance analysis allowed us to interpret which lagged inputs most significantly contributed to model output. This made XGBoost especially valuable in identifying leading indicators of future loss events—such as rapid changes in preceding days’ values—and in offering transparency in model behavior.

\begin{figure}[H]
    \centering
    \includegraphics[width=0.95\textwidth]{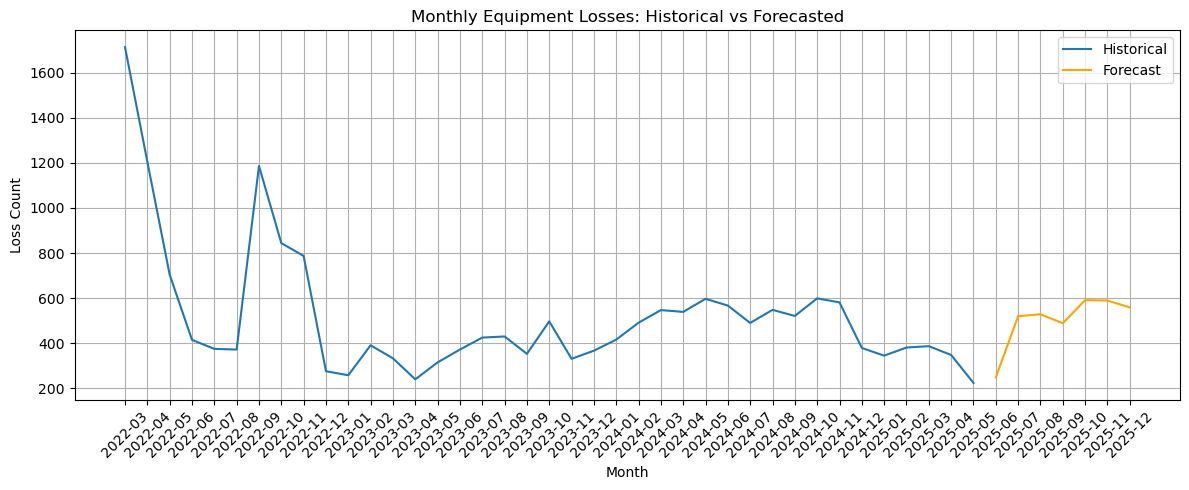}
    \caption{Monthly Equipment Loss Forecast (All Types) via XGBoost}
\end{figure}

\begin{figure}[H]
    \centering
    \includegraphics[width=0.95\textwidth]{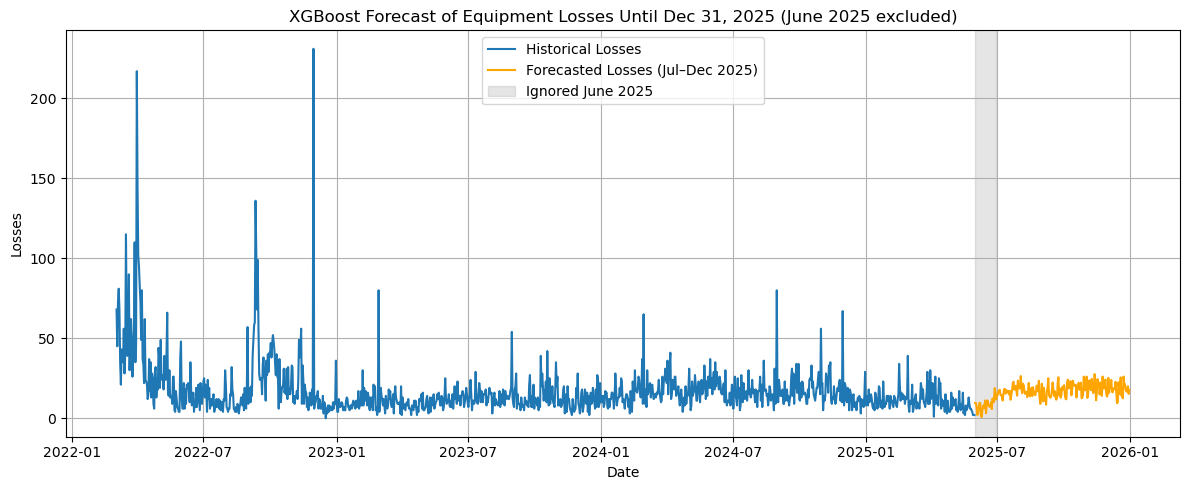}
    \caption{XGBoost Forecast of Daily Equipment Losses (Jul--Dec 2025), June Omitted}
\end{figure}

\begin{figure}[H]
    \centering
    \includegraphics[width=0.95\textwidth]{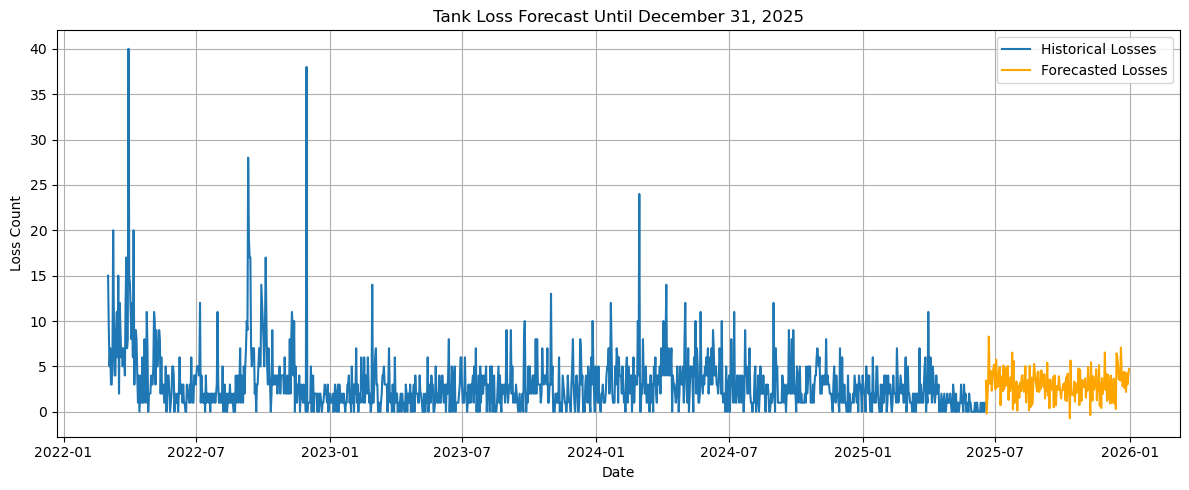}
    \caption{XGBoost Forecast: Daily Tank Losses Only (Jul--Dec 2025)}
\end{figure}

XGBoost, using calendar and time-series features, projects increased losses from July to October 2025, possibly tied to seasonal offensives or replenishment cycles. Such modeling approaches have previously been used to track operational dynamics in political conflict datasets (\cite{raleigh2010acled}.) The daily-level forecast suggests a return to mid-2024 loss levels. Tank-specific results imply sustained attrition at 3--5 vehicles per day, reflecting both continued engagement and the depletion of newer tank platforms in favor of IFVs and older Soviet stock.
Figures 4 through 6 present gradient-boosted forecasts of Russian equipment losses generated using the XGBoost algorithm. Figure 4 depicts the monthly aggregate of all equipment losses (excluding June 2025), revealing a sharp decline following early 2022 peaks, followed by moderate stabilization between 2023 and 2025. The forecast suggests a slight upward trend through the remainder of 2025. Figure 5 disaggregates this into daily resolution for all equipment types, again excluding June 2025 due to inconsistent reporting. The daily series shows high-frequency volatility in 2022, gradually tapering to a more stable baseline in 2024–2025. Forecasted values for July–December 2025 indicate an increase in variance but maintain a relatively consistent mean. Figure 6 isolates daily tank losses, highlighting a similar pattern of sharp early peaks, long-term decline, and model-predicted stabilization between 4–8 losses per day through the end of 2025. Collectively, these visualizations demonstrate XGBoost’s capacity to detect nonlinear seasonality and recover from missing intervals (e.g., June 2025) while offering high-resolution forecasts under complex temporal dynamics.

\subsection*{LSTM Forecasts (w/ Weekday Metadata)}
We employed LSTM due to its proven ability to model long-term dependencies in sequential data—a critical strength when forecasting battlefield losses shaped by historical momentum, operational tempo, and lagged effects of combat engagements. LSTM networks incorporate gated memory units that allow the model to retain relevant information from many prior time steps, enabling it to capture complex, nonlinear behaviors in the data. In contrast to models like ARIMA and Prophet, which are constrained by explicit trend or seasonality assumptions, LSTM learns patterns directly from data, making it more flexible when modeling sudden escalations or unpredictable fluctuations—such as those observed during major offensives in regions like Bakhmut or Avdiivka.

\begin{figure}[H]
    \centering
    \includegraphics[width=0.95\textwidth]{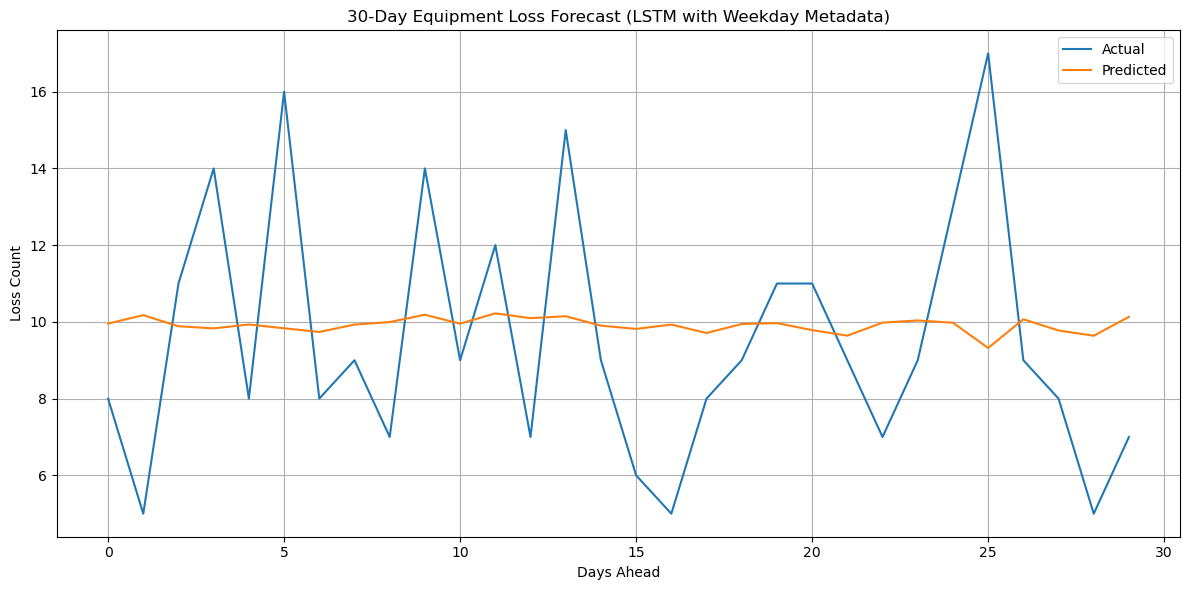}
    \caption{30-Day Equipment Loss Forecast using LSTM with Weekday Metadata}
\end{figure}

\begin{figure}[H]
    \centering
    \includegraphics[width=0.95\textwidth]{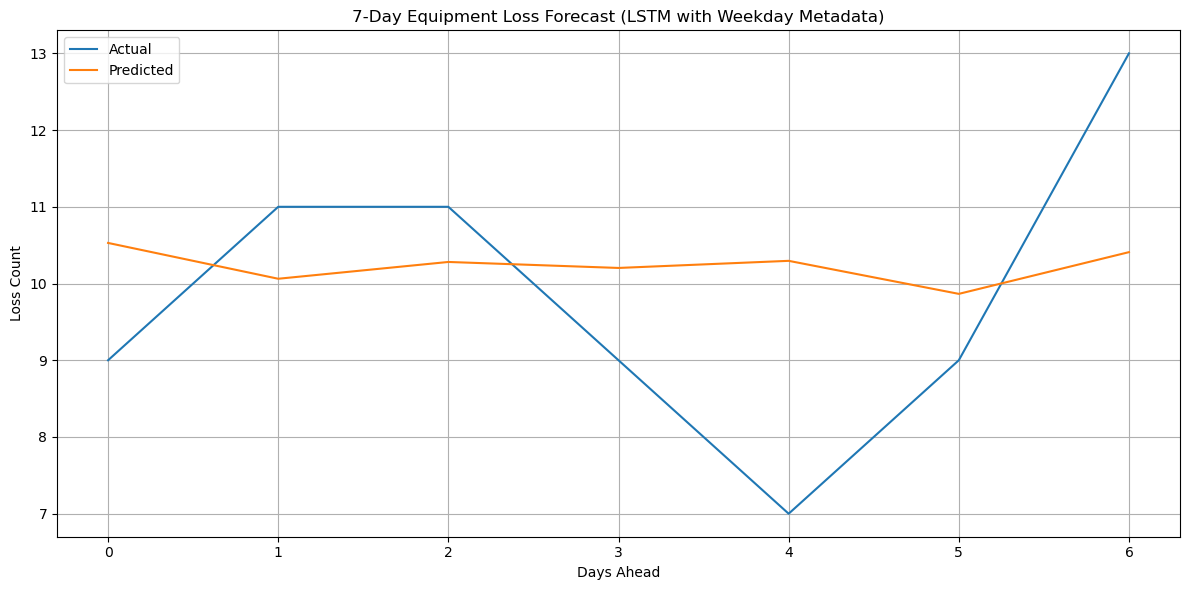}
    \caption{7-Day Equipment Loss Forecast (LSTM)}
\end{figure}

\begin{figure}[H]
    \centering
    \includegraphics[width=0.95\textwidth]{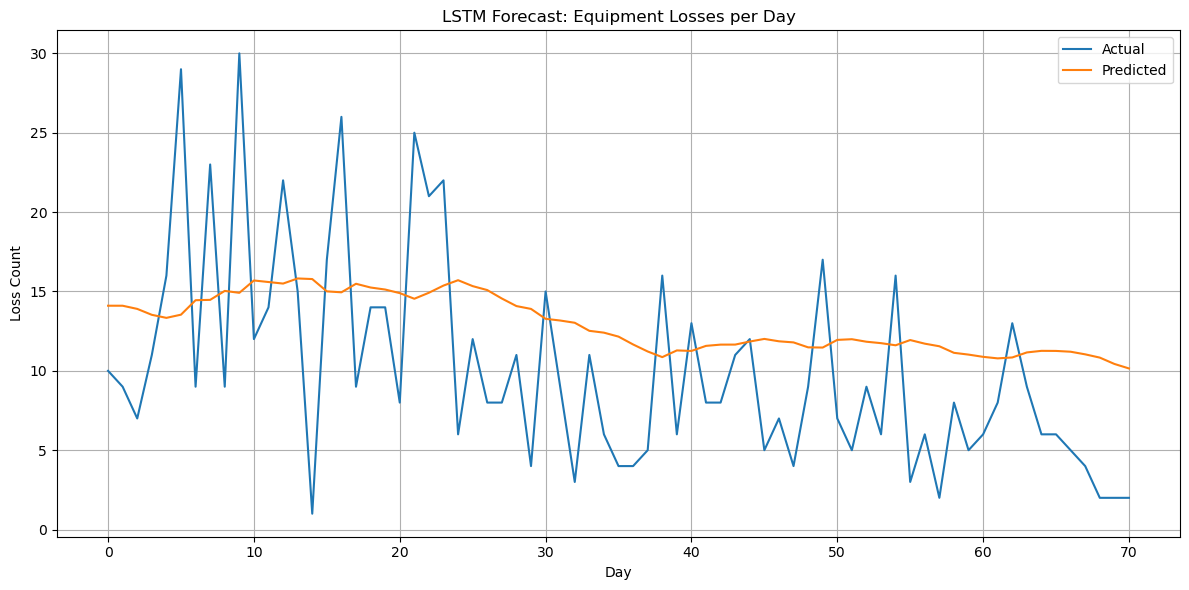}
    \caption{70-Day Daily Loss Forecast: LSTM Model with Seasonal Memory}
\end{figure}

LSTM models capture weekday effects in OSINT confirmation, such as weekend lulls and Monday upload surges. Though capable of modeling seasonality, LSTM forecasts tend to underreact to escalation events and produce smoothed predictions. Overall, the LSTM predicts stable low-loss regimes, suggesting that current visibility of confirmed losses is primarily shaped by reporting practices rather than combat shocks. Machine learning models like LSTM have been increasingly applied in conflict prediction tasks to handle nonlinearities in reporting dynamics (\cite{muchlinski2021machine}.)

Figures 7 through 9 illustrate short- and medium-range forecasts of daily Russian equipment losses using Long Short-Term Memory (LSTM) models. Figure 7 presents a 30-day forecast incorporating weekday metadata, where the model generates a smooth predictive baseline around 10 losses per day, despite high variance in the actual data. While the LSTM fails to capture short-term spikes and dips, it does preserve overall trend structure. Figure 8 refines this to a 7-day forecast, showing closer alignment between observed and predicted values, especially in periods of lower volatility—highlighting the LSTM’s strength in capturing near-term patterns. Finally, Figure 9 extends the forecasting horizon to 70 days using a seasonal LSTM variant. Here, the model exhibits better tracking of broader cyclical trends but continues to underestimate short-term fluctuations and local extrema. Across all three figures, the LSTM model demonstrates utility in smoothing volatility and projecting stable trajectories, though its deterministic predictions often lag behind abrupt shifts in battlefield tempo.
\subsection*{Prophet Forecasts (Facebook Prophet)}
We employed Prophet for its robustness in modeling irregular, sparse, and seasonally structured data—characteristics frequently observed in OSINT-based military time series.Prophet is especially useful for automatically detecting trend changes, such as escalation or de-escalation in battlefield activity, without requiring extensive manual tuning. Its additive decomposition into trend, seasonality, and holiday effects aligns well with the hypothesis that certain military surges or pauses may follow strategic, political, or logistical cycles. In our case, Prophet’s robustness to missing data and changepoint flexibility made it a valuable model for interpreting the temporal structure of equipment losses while accounting for noise and reporting delays.

\begin{figure}[H]
    \centering
    \includegraphics[width=0.95\textwidth]{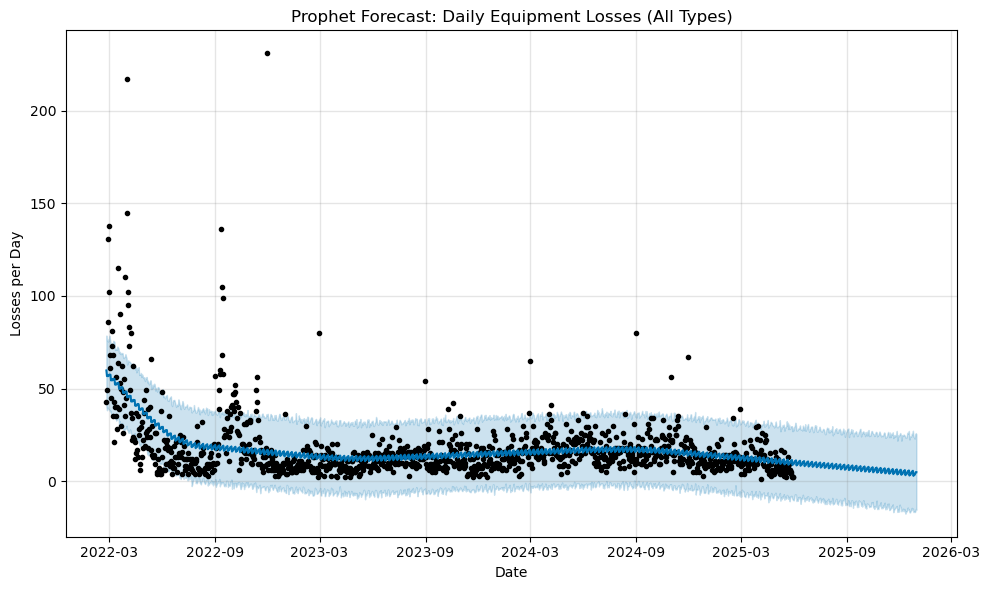}
    \caption{Prophet Model: Forecast of All Equipment Losses (2022--2026)}
\end{figure}

\begin{figure}[H]
    \centering
    \includegraphics[width=0.95\textwidth]{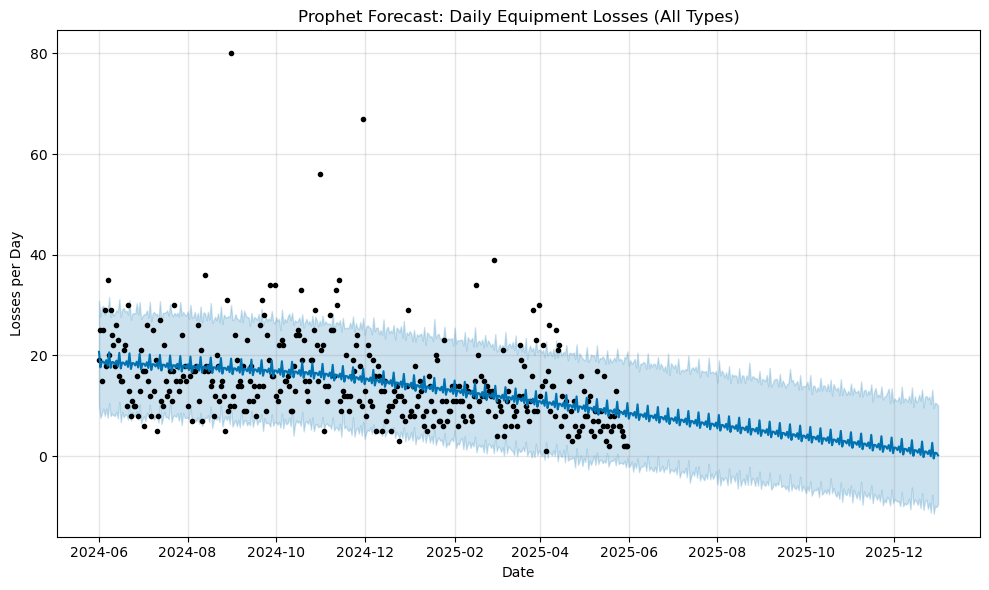}
    \caption{Prophet Model Zoomed on 2024--2025 Loss Forecast}
\end{figure}

Prophet decomposes trends and seasonality, forecasting a slow decline in daily confirmed losses through 2025. The trend reflects logarithmic decay---initially high in early 2022, then stabilizing in 2023--2024, with gradual tapering projected through 2025. This is consistent with the depletion of armored platforms and ongoing adaptation of Russian logistics under sanctions.
Figures 10 and 11 present daily equipment loss forecasts generated using the Prophet time series model, capturing trends from 2022 through early 2026. In Figure 10, the full series displays a steep early decline following the 2022 invasion peak, followed by a relatively stable but noisy plateau through 2023 and 2024. The model projects a gradual long-term decrease in average daily losses, with the central forecast (dark blue line) flanked by a widening 95\% uncertainty interval (light blue ribbon). Figure 11 zooms in on the 2024–2025 segment, revealing Prophet’s internal seasonal structure, which captures weekly loss seasonality and a declining baseline trend. Despite some high outliers in the raw data, the model remains conservative, centering forecasts around 15–20 losses per day and gradually trending downward. These results highlight Prophet’s strength in decomposing trend and seasonality while offering probabilistic intervals for medium-horizon military forecasting under irregular reporting conditions.
\section*{Limitations of OSINT-Based Loss Data (WarSpotting)}

While OSINT-based databases such as WarSpotting provide invaluable transparency into battlefield developments, they carry significant limitations that must be considered when interpreting model results. These limitations affect data completeness, timeliness, classification accuracy, and spatial consistency. Acknowledging these constraints is essential for contextualizing forecasts and avoiding overconfident interpretations of attrition dynamics.

\paragraph{}
WarSpotting relies exclusively on visually confirmed equipment losses, which introduces structural undercounting. This confirmation bias arises because only those losses captured in publicly accessible visual media—such as drone footage, field photography, or video reports shared via Telegram, YouTube, and Twitter—are included in the dataset. As a result, losses that occur without visual documentation, particularly in media-restricted areas such as deep frontlines, Russian-held territories, or during blackouts, remain uncounted. This makes the dataset inherently conservative, capturing a lower bound of total losses rather than a full accounting.

\paragraph{}
The date of visual confirmation often lags behind the actual loss event, sometimes by weeks. This introduces a temporal distortion in the dataset, complicating efforts to analyze or forecast trends in near-real-time. In high-activity zones, a tank may be destroyed during combat but not confirmed via OSINT sources until weeks later—particularly if the terrain is contested or the imagery requires vetting. This asynchronous reporting can cause spikes in model outputs that reflect delayed data rather than true battlefield dynamics, especially if models assume losses are recorded contemporaneously.

\paragraph{}
The WarSpotting data exhibits spatial skew: regions with high drone surveillance, embedded journalism, or pro-Ukrainian channels are more heavily represented. Areas such as Donetsk and Zaporizhzhia, where Ukrainian drone coverage and local OSINT contributors are more active, dominate the dataset. In contrast, less media-visible regions like rural areas of Luhansk or deep Russian-held zones receive minimal documentation. Moreover, visually distinctive or politically symbolic systems—such as T-90 tanks or S-400 systems—are overrepresented relative to less notable platforms like logistical vehicles, electronic warfare trucks, or engineering assets. This introduces significant sampling bias in loss profiles by region and equipment category.

\paragraph{}
Equipment types are occasionally misclassified due to low-resolution imagery or partially destroyed platforms. When a vehicle is captured in damaged or obstructed condition, analysts may misidentify it—for instance, confusing a BTR-80 with a BTR-82A, or a T-72B3 with a T-80BV. These errors can propagate into aggregate category-level statistics, especially when vehicles share chassis or silhouette characteristics. As a result, forecast models that rely on accurate classification counts—such as those modeling IFV or tank losses—may inherit structural uncertainty from the underlying visual evidence.

\paragraph{}
The dataset provides a view of losses but not of deployed or remaining inventories. This “denominator problem” limits the ability to interpret loss rates in proportion to total forces. While models can project the frequency and accumulation of confirmed losses, they cannot determine the relative degradation of a given battalion, division, or theater-wide capability. Without access to inventory baselines—such as initial pre-war stockpiles, ongoing reinforcements, or production outputs—any inferred attrition rate remains speculative. This problem is especially acute in forecasting Russian force degradation, where state secrecy obscures order-of-battle details.

\paragraph{}
Changes in Telegram moderation, Ukrainian or Russian censorship, or the deplatforming of key OSINT actors can cause data availability to fluctuate. WarSpotting’s dependence on open platforms makes it vulnerable to shifts in digital policy and enforcement. For example, if Telegram cracks down on pro-Ukrainian military channels or if Russia successfully censors specific war content, the volume and variety of confirmed loss data may drop. Such platform volatility can artificially suppress new entries and mislead time series models into perceiving declining loss rates that are in fact artifacts of reduced access, not improved battlefield survivability.

\paragraph{}
While WarSpotting remains one of the most rigorous public-facing OSINT repositories, its visual-confirmation methodology inherently prioritizes documentation over completeness. The result is a dataset that is immensely useful but fundamentally partial—its counts should always be interpreted as minimum confirmed losses, not exhaustive tallies. Further, although deep learning models such as LSTM and TCN perform well in projecting future values, their internal representations are opaque, and they can reflect data artifacts such as reporting lags or platform bias without transparency. This introduces limits to interpretability and trust, particularly in high-stakes applications like military planning or diplomatic forecasting.

\paragraph{}
Nonetheless, the act of performing analysis with such datasets is not without precedent. Our symbolic predictor draws on curated OSINT streams and battlefield observations, such as those compiled by SIPRI and similar monitors (\cite{boulanin2020ai}). While imperfect, these datasets allow researchers and analysts to model conflict evolution with greater granularity than state-level reporting or classified briefings alone. Still, all forecasts derived from WarSpotting must be accompanied by careful methodological disclaimers and a recognition of what remains unseen.

\section*{Conclusion}

All models agree on the broad trend of stabilized or declining Russian equipment loss rates through late 2025. The TCN and LSTM models suggest stagnation in tank availability or verification throughput, while Prophet reinforces the long-term attrition curve consistent with logistical depletion. XGBoost forecasts, in contrast, anticipate a modest uptick in losses during the second half of 2025—possibly reflecting lagged OSINT updates or seasonal operational surges.

Among the methods tested, TCN and LSTM offered the most temporally consistent forecasts at the daily level, while Prophet provided smoother strategic trends. ARIMA served as a conservative statistical baseline, particularly useful when excluding anomalous reporting periods. The variety of temporal assumptions embedded across models reinforces the need for multi-perspective ensemble approaches when forecasting under uncertainty. Such ensemble approaches have been advocated in the conflict forecasting literature \cite{ward2013system}.

Future work could integrate geospatial predictors, unit-level loss affiliations, and event-based triggers—such as the start dates of major offensives—to enhance both tactical interpretability and operational relevance. Incorporating spatial variables would allow for more granular mapping of loss patterns and their correlation with geographic features, while unit-level affiliations could reveal differential vulnerability or resilience among specific formations. Event-based triggers, in turn, could capture shifts in loss dynamics linked to strategic or operational milestones, enabling models to better reflect the temporal rhythms of conflict. Together, these enhancements would create a more nuanced analytical framework, improving both forecasting accuracy and practical utility for operational planning.
\section{Acknowledgments}
\section{Declaration Of Competing Interests}
The author report there are no competing interests to declare.
\printbibliography
\end{document}